%% file: iclr2025_conference.tex
\documentclass{article} %
\usepackage{iclr2025_conference,times}

\input{math_commands.tex}

\usepackage{epigraph}
\usepackage{graphicx}
\usepackage[utf8]{inputenc} %
\usepackage[T1]{fontenc}    %
\usepackage{hyperref}       %
\usepackage{url}            %
\usepackage{booktabs}       %
\usepackage{amsfonts}       %
\usepackage{nicefrac}       %
\usepackage{microtype}      %
\usepackage{floatrow}
\usepackage{subcaption}
\usepackage{multirow}
\usepackage{listings}
\usepackage{amsmath}
\usepackage{makecell}
\newcommand{\methodname}{\texttt{{{PROVE}}}}
\newcommand{\dsize}{10.5k}
\newcommand{\isize}{5k}

\title{Trust but Verify: \\ Programmatic VLM Evaluation in the Wild}

\author{Viraj Prabhu, Senthil Purushwalkam, An Yan, Caiming Xiong \& Ran Xu \\
Salesforce AI Research\\
{\small\texttt{\{viraj.prabhu,spurushwalkam,an.yan,cxiong,xurantju\}@salesforce.com}
}}

\iclrfinalcopy %
\begin{document}

\maketitle

\vspace{-5pt}
\begin{abstract}
   Vision-Language Models (VLMs) often generate plausible but incorrect responses to visual queries. However, reliably quantifying the effect of such hallucinations in free-form responses to open-ended queries is challenging as it requires visually verifying each claim within the response. We propose \underline{Pro}grammatic \underline{V}LM \underline{E}valuation (\methodname), a new benchmarking paradigm for evaluating VLM responses to open-ended queries. To construct \methodname, we provide a large language model (LLM) with a high-fidelity scene-graph representation constructed from a hyper-detailed image caption, and prompt it to generate diverse question-answer (QA) pairs, as well as programs that can be executed over the scene graph object to \emph{verify} each QA pair. We thus construct a benchmark of \dsize~challenging but visually grounded QA pairs. Next, to evaluate free-form model responses to queries in \methodname, we propose a \emph{programmatic} evaluation strategy that measures both the helpfulness and truthfulness of a response within a unified scene graph-based framework. We benchmark the helpfulness-truthfulness trade-offs of a range of VLMs on \methodname, finding that very few are in-fact able to achieve a good balance between the two. \\{\centering Project page: \url{https://prove-explorer.netlify.app/}}.
\end{abstract}
\vspace{-5pt}

\input{sections/1_intro}
\input{sections/2_relwork}
\input{sections/3_method}
\input{sections/4_experiments}
\input{sections/5_discussion}

\bibliography{iclr2025_conference}
\bibliographystyle{iclr2025_conference}

\appendix
\section{Appendix}

\input{sections/X_suppl}

\end{document}

%% file: math_commands.tex
\usepackage{amsmath,amsfonts,bm}

\def\eqref#1{equation~\ref{#1}}

\def\1{\bm{1}}

\DeclareMathAlphabet{\mathsfit}{\encodingdefault}{\sfdefault}{m}{sl}
\SetMathAlphabet{\mathsfit}{bold}{\encodingdefault}{\sfdefault}{bx}{n}

%% file: sections/1_intro.tex
\section{Introduction}
\label{sec:intro}

Vision-language models (VLMs) have emerged as an effective solution for generating responses to queries about visual content. However, despite impressive progress (and much like their LLM-counterparts), VLMs are still known to hallucinate -- to generate plausible but incorrect answers that are either inconsistent or unverifiable against the provided visual context~\footnote{A few LLM-focused works also consider responses that contradict \emph{world knowledge} as hallucinations, but we exclude these from our scope.}. This crucial shortcoming has the potential to erode trust in such systems and has already begun to attract significant research~\citep{yu2024rlhf,liu2023hallusionbench,gunjal2024detecting,huang2024opera} and regulatory~\citep{biden2023executive} interest, particularly as using such models as the ``foundation'' of various high-stakes applications becomes imminent~\citep{bommasani2021opportunities}. 

This has led to a flurry of research on \emph{reliably} benchmarking VLM performance~\citep{liu2024survey}, by measuring not just the helpfulness but also the \emph{truthfulness} of their responses. Existing benchmarks fall into two categories -- \emph{discriminative}~\citep{hu2023ciem,lovenia2023negative,li2023evaluating}, which evaluate the model's responses to close-ended, existence-based queries (``Is there a man in this image?''), and \emph{generative}~\citep{rohrbach2018object,sun2023aligning,liu2023mitigating,liu2023hallusionbench,gunjal2024detecting}, which evaluate responses to free-form, open-ended questions (``Describe this image.''). While discriminative benchmarks ease evaluation, they do not realistically simulate in-the-wild usage. On the other hand, generative benchmarks, while realistic, are \emph{extremely} challenging to reliably evaluate, as they require verifying both that the model response fully answers the question (\emph{i.e.} is helpful) and does not make any false claims (\emph{i.e.} is truthful).

\begin{figure}[t]
    \centering
    \includegraphics[width=1.0\linewidth]{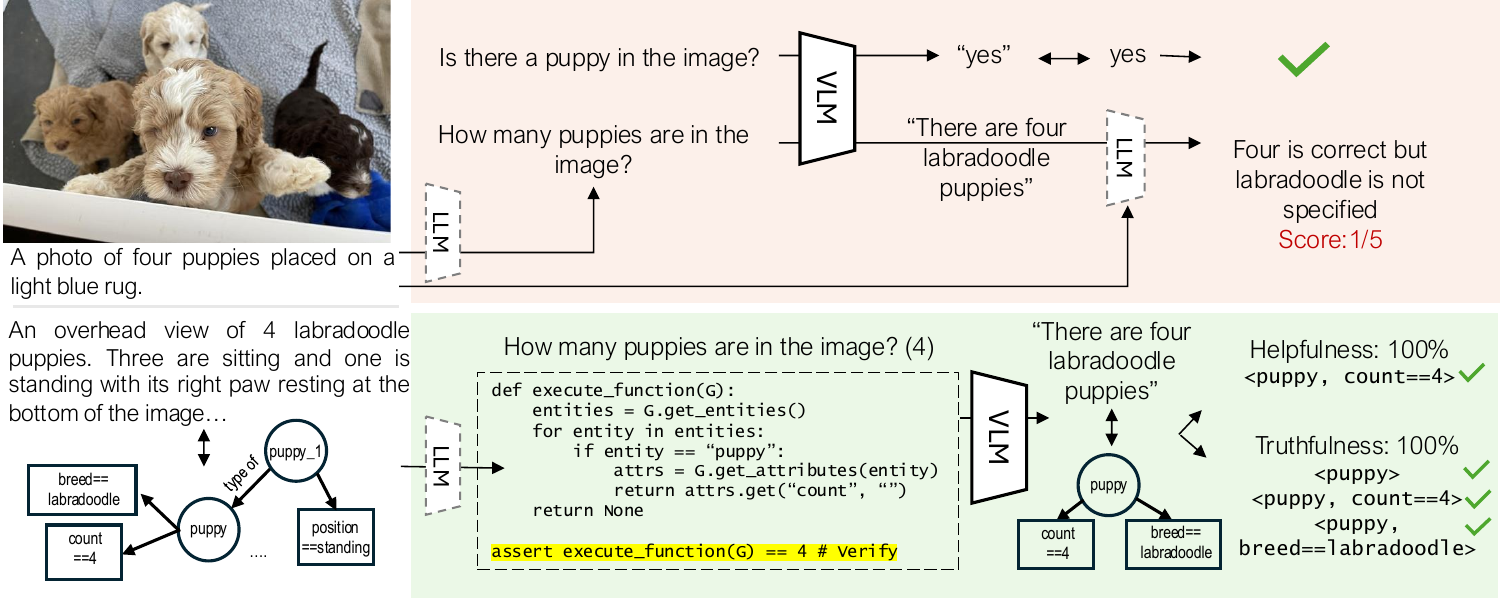}
    \caption{\textbf{Top.} Existing VLM benchmarks either limit query-types to easy-to-evaluate but restrictive binary questions, or use external LLMs to generate open-ended questions (without verifying their validity) and score answers (often without complete image context or a clear scoring rubric). \textbf{Bottom.} We propose \methodname, a new benchmark that constructs high-fidelity scene-graph representations from hyper-detailed image captions, that are queried via an LLM-generated program to verify a free-form generated question-answer pair. At test-time, we perform an interpretable programmatic evaluation of the helpfulness and truthfulness of free-form VLM responses by comparing scene-graphs.}
    \label{fig:teaser}
\end{figure}

Evaluating such free-form responses typically relies on external models (usually, a proprietary LLM) to score responses given some image context (typically ground-truth annotations). However, we find that in several such benchmarks, the context provided is completely insufficient to judge if the response contains hallucinations. Consider Fig.~\ref{fig:teaser}: a VLM may respond to the query ``How many puppies are in the image?'' (correct answer = ``four''), with ``There are four labradoodle puppies''. Evaluating the truthfulness of this statement requires verifying multiple claims about the puppies (\texttt{<count == four>} and \texttt{<breed == labradoodle>}); however, an LLM judge provided only with a brief image caption as context (``four puppies placed on a light blue rug'') will be unable to do so! Further, the absence of a clear scoring rubric coupled with the sensitivity of LLMs to minor prompt differences often leads to inconsistent and arbitrary scores in such cases. In Fig.~\ref{fig:comparison}, we provide real examples from existing benchmarks that illustrate these challenges.

We propose Programmatic VLM Evaluation (\methodname), a new evaluation paradigm that performs reliable and interpretable \emph{programmatic} evaluation of free-form VLM responses to challenging, diverse, and grounded questions. To build this dataset, we first use hyper-detailed image captions to construct a high-recall scene graph image representation. We then use an LLM to generate a diverse set of open-ended question-answer (QA) pairs along with  accompanying \emph{verification programs}.
While the QA pairs are meant to test a range of model capabilities under real-world use, the verification programs can be executed over a given scene graph object to verify the correctness and groundedness of its corresponding QA pair. We only retain the QA pairs that we can programmatically verify and construct a benchmark of \dsize~ diverse and challenging examples which are visually grounded by design, that we can use to reliably benchmark VLM responses.

Next, we benchmark VLM responses to queries in \methodname~by comparing scene graph representations. First, we measure the helpfulness of a response by computing its scene graph-based \emph{recall} against the ground truth answer. Next, we measure response truthfulness as its scene graph-based \emph{precision} against both the scene-graph constructed from the full caption or the image itself. We benchmark a range of VLM responses using this approach, and study their respective trade-offs between helpfulness and truthfulness. Our findings suggest that much of the recent progress in training ``better'' VLMs also translate to improved helpfulness on our benchmark, but not necessarily to higher truthfulness.

%% file: sections/2_relwork.tex
\section{Related work}
\label{sec:relwork}

\begin{figure}[t]
    \centering
    \includegraphics[width=\linewidth]{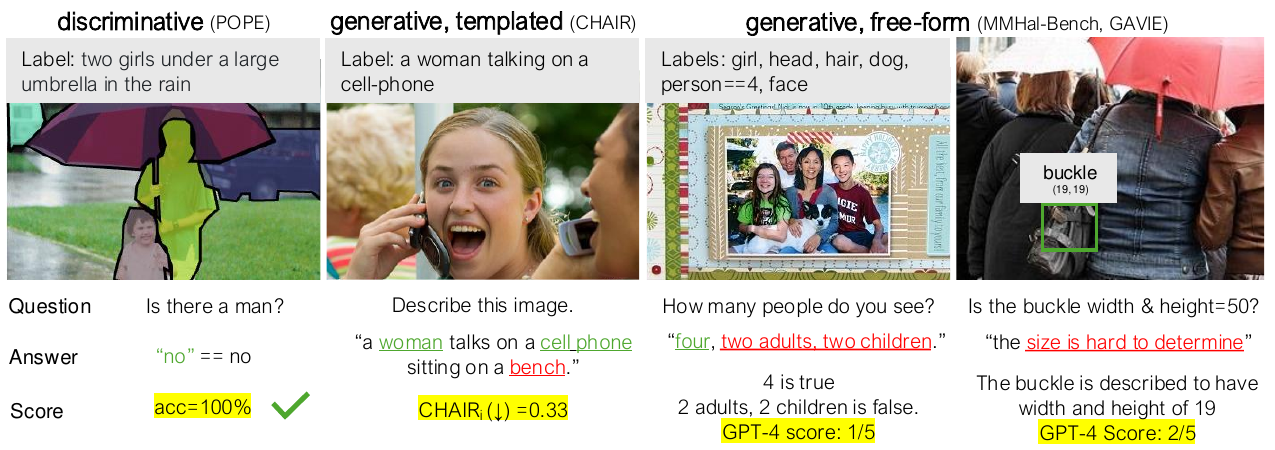}
    \caption{\textbf{Top.} Existing VLM hallucination evaluation benchmarks either measure VLM performance on object existence queries (``discriminative''~\citep{li2023evaluating}) or object precision/recall in generated image captions (``generative, templated''~\citep{rohrbach2018object}), neither of which realistically simulate in-the-wild usage. Some recent benchmarks contain open-ended queries (``generative, free-form''~\citep{sun2023aligning}), which are more realistic but also harder to both generate (\emph{e.g.} see unnatural QA-pair from GAVIE~\citep{liu2023mitigating} -- first from right), and evaluate with an LLM-as-judge (\emph{e.g.} see GPT-4 penalizing a correct response that includes details absent from the ground truth in MMHal-Bench~\citep{sun2023aligning} -- second from right). \textbf{Bottom.} We propose \methodname~, a benchmark of challenging but verifiable open-ended questions that we use to jointly evaluate both the truthfulness and helpfulness of free-form model responses.}
    \label{fig:comparison}
\end{figure}

\noindent\textbf{Benchmarking VLM hallucination.} Existing benchmarks fall into one of two groups (see Fig.~\ref{fig:comparison}):

\noindent\textbf{$\triangleright$ Discriminative benchmarks} generate a series of binary questions to verify the presence (or absence) of various entities (or distractors) in the image. Early benchmarks like POPE~\citep{li2023evaluating} limited their scope to object entities annotated by humans or external off-the-shelf models~\citep{zou2024segment}, whereas follow-up works additionally evaluate responses to \emph{negative presence} queries~\citep{lovenia2023negative}, which stress-test the model's abstention capabilities on questions about entities \emph{absent} from the image, or use an LLM to generate a broader range of existence-based questions covering objects and their attributes~\citep{hu2023ciem}. However, while the binary questions that typify such benchmarks simplify evaluation, they do not realistically simulate in-the-wild use.

\noindent\textbf{$\triangleright$ Generative benchmarks} instead evaluate model hallucinations in response to free-form questions. CHAIR~\citep{rohrbach2018object} measures the precision and recall of entities mentioned in a generated image description against the ground truth. HaELM~\citep{wang2023evaluation} additionally uses a large language model (LLM) to judge generations, whereas M-HalDetect~\citep{gunjal2024detecting} has humans annotate hallucinations in model generated descriptions are used to train a predictive model. Recently, AMBER~\citep{wang2023llm} combines a POPE style evaluation with a generative evaluation over an open-ended split. While these benchmarks are indeed more realistic, they still restrict the query instruction to image captioning-style templates (``Describe this image in detail.''). 

Most recently, a few benchmarks with truly open-ended queries have been proposed~\citep{sun2023aligning,liu2023hallusionbench,jing2023faithscore,liu2023mitigating}, which either hand-design or use an LLM to generate free-form questions, and use external models to judge the corresponding responses. However, these too have limitations: MMHal~\citep{sun2023aligning} and HallusionBench~\citep{liu2023hallusionbench} rely on a series of off-the-shelf models at various stages which introduce noise (see Fig.~\ref{fig:comparison}, col 3). GAVIE's~\citep{liu2023mitigating} reliance on dense captions and bounding boxes leads to a majority of questions querying localized image regions and spatial relationships, many of which have unnatural-sounding responses (\emph{eg.} mentioning image coordinates, see Fig.~\ref{fig:comparison}, col 4). Finally, GPT-4-based evaluation is both expensive and inherits the model's own limitations. 

\noindent\textbf{Understanding and mitigating VLM hallucination.} Several works have sought to better understand \emph{why} VLMs hallucinate. One prevalent theory is the model learning spurious correlations between the input and the output: either due to overly strong text priors learned by the LLM backbone~\citep{huang2024opera,leng2024mitigating}, or due to distilling synthetic outputs generated by stronger models (such as GPT-4V) that may themselves contain confabulation~\citep{liu2024llava}. This is often exacerbated by the predominant training recipe~\citep{liu2024llava,abdin2024phi} that learns a shallow projection from the visual input to the text embedding space which limits the expressivity of the model to learn visually grounded representations.

Recent work has proposed training-based and training-free strategies for mitigating hallucinations. The former involves finetuning~\citep{liu2023mitigating} or preference optimization~\citep{yu2024rlhf,sun2023aligning} of ``preferred'' ground truth responses against dis-preferred synthetically generated ``hallucinations''. Training-free methods instead focus on specialized decoding strategies~\citep{huang2024opera,kim2024if,leng2024mitigating} that seek to correct for potential statistical bias that may lead to hallucination. 

However, developing better understanding and mitigation strategies are both contingent on the availability of reliable evaluation benchmarks. In this work, we introduce such a benchmark of challenging but verifiable open-ended visual questions that we use to jointly evaluate both the truthfulness and helpfulness of free-form model responses.

%% file: sections/3_method.tex
\newcommand*{\question}{\ensuremath{\mathcal{Q}}}
\newcommand*{\image}{\ensuremath{\mathcal{I}}}
\newcommand*{\imagecaption}{\ensuremath{\mathcal{C}}}
\newcommand*{\answer}{\ensuremath{\mathcal{A}}}
\newcommand*{\scenegraph}{\ensuremath{\mathrm{g}}}
\newcommand*{\help}{\ensuremath{\mathsf{hscore}}}
\newcommand*{\truth}{\ensuremath{\mathsf{tscore}}}
\newcommand*{\response}{\ensuremath{\hat{\mathcal{A}}}}
\newcommand*{\model}{\ensuremath{\mathrm{m}_{\theta}}}
\newcommand*{\entity}{\texttt{<entity>}}
\newcommand*{\attribute}{\texttt{<entity, attribute>}}
\newcommand*{\relationship}{\texttt{<entity\_1, attribute, entity\_2>}}

\section{Approach}
\label{sec:method}

\begin{figure}[t]
    \centering
    \includegraphics[width=0.99\linewidth]{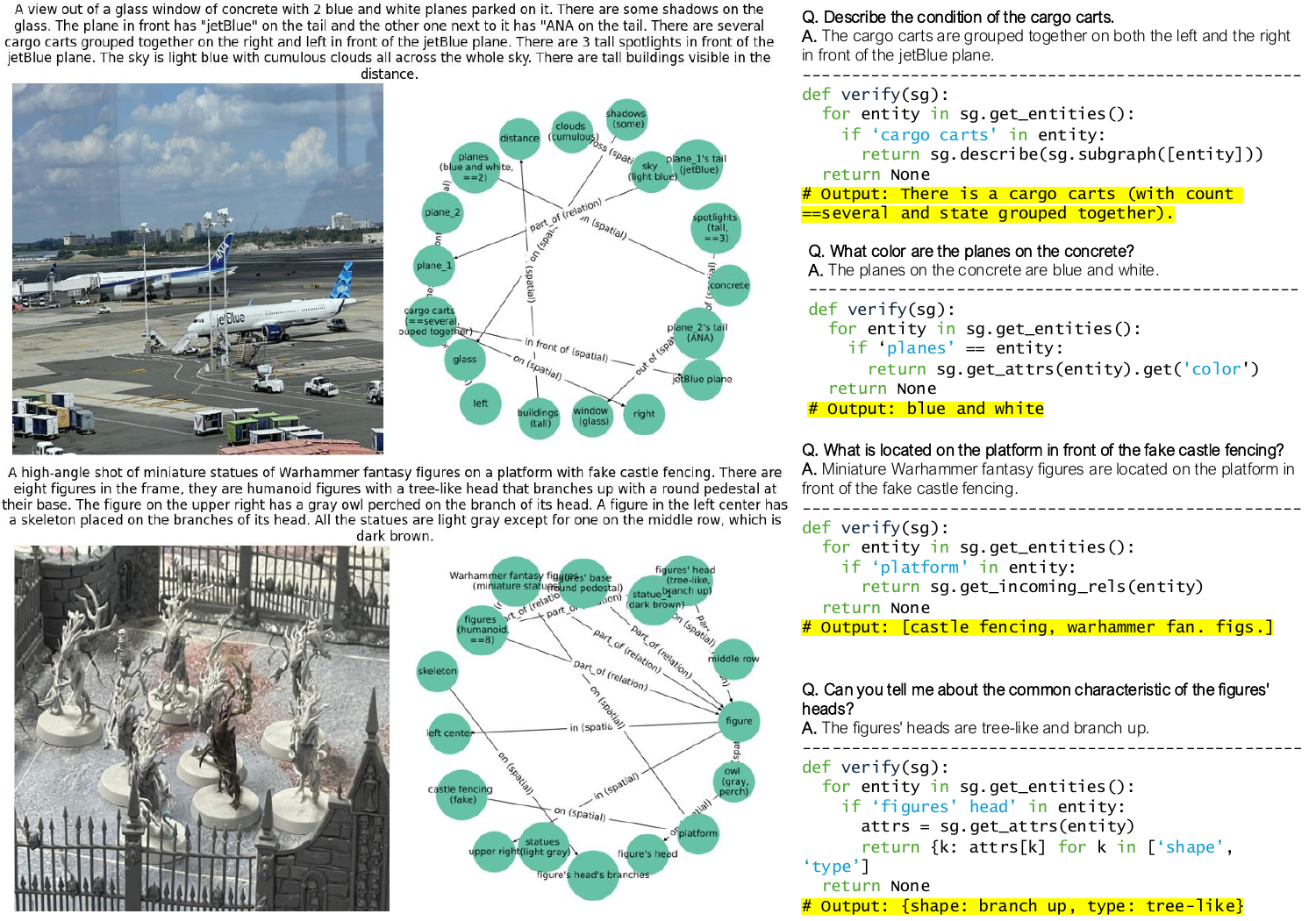}
    \caption{\textbf{The \methodname~ dataset}. For each image-caption pair, we generate a high-fidelity scene graph representation with which we prompt an LLM to generate challenging QA pairs and their verification programs. We only retain QA pairs that we can programmatically verify, ensuring diverse but reliable evaluation data that is grounded by design.}
    \label{fig:examples}
\end{figure}

Vision-language models are trained to respond to a question \question~ about an image \image~ with a ground-truth answer \answer. Let \model(.) denote a VLM model trained on a large dataset of such (\image, \question, \answer) triplets. At test time, we wish to evaluate the model response \response=\model(\question, \image). Specifically, while prior work typically evaluates either the response's correctness (is \response$=$\answer) or truthfulness (is $p$(\response|\image) $>$ threshold), we propose a unified framework that jointly evaluates both.

\subsection{Generating verifiable Visual Question-Answers}
\label{sec:build}

To build \methodname~, we first download image-caption pairs (\image, \imagecaption) from the test set of the recently proposed DOCCI~\citep{onoe2024docci} dataset, containing 5k manually curated images with comprehensive human-annotated descriptions. DOCCI is particularly well-suited for VLM evaluation because: i) its captions are extremely detailed, with a higher median caption length than competing datasets, which correlates with high image recall ii) its comprehensive and rigorous 3-stage human annotation protocol leads to high-fidelity captions that are suitable to test a range of image understanding challenges including spatial reasoning, counting, text rendering, and compositionality, and iii) its images are newly curated and so more likely to be truly held-out for existing models.

\noindent\textbf{Building a robust scene-graph representation.} Scene-graphs are comprised of entity (\entity), attribute (\attribute), and relationship (\relationship) tuples that describe a scene. We use the tuples included with the DOCCI test set that were automatically extracted from its captions using an LLM~\citep{chodavidsonian}, and use it to construct a scene graph representation \scenegraph(\imagecaption) as a directed graph with attributed entities as nodes and relationships as edges. The scene graph is implemented as a Python class with methods to query the graph for its entities, attributes, and relationships, as well as to extract and describe subgraphs in natural language (full API in Lst.~\ref{lst:python}). 

\noindent\textbf{Generating open-ended questions with verifiable answers.} Next, for each image, we prompt a pre-trained LLM to generate 10-15 challenging, diverse, and unambiguous question-answer (QA) pairs given a caption and scene graph, along with an accompanying Python program that accepts the scene graph as input and can be executed to verify the generated QA pair~\citep{gupta2023visual,suris2023vipergpt}. We include a few in-context examples of such scene-graph and QA+program input/output pairs in the prompt (see Fig.~\ref{fig:prompt1}). We repeat this procedure to generate a large dataset of open-ended image+QA pairs $\{(\mathcal{I}_i, \mathcal{Q}_i, \mathcal{A}_i)\}_{i=1}^N$ and their verification programs.

\noindent\textbf{Filtering QA pairs.} Next, we perform two rounds of filtering: 

\textbf{1. Programmatic:} First, we execute the generated program with the scene graph as input to verify the QA pair. We discard pairs for which the program either fails or returns an answer that is semantically different~\citep{reimers2019sentence} from the ground truth answer.

\noindent\textbf{2. Text-based:} Next, we perform a few additional post-processing steps to exclude low-quality QA pairs which are i) trivial, ungrammatical, ambiguous, or incomplete (using an LLM, see Fig.~\ref{fig:prompt2}), ii) not entailed by the image (using a visual entailment model~\citep{cai2019once}), iii) include one or more words from a manually curated list of taboo words that we find to result in low-quality questions, or iv) semantic duplicates for the same image (using SemDeDup~\citep{abbas2023semdedup}). Our final dataset after filtering contains ~\dsize~ high-quality visual question answers -- see Fig.~\ref{fig:examples}. 

\noindent\textbf{Dataset statistics.} We now present some statistics about \methodname, which comprises of \dsize~QA pairs generated from \isize~image-caption pairs from the DOCCI test set. These are obtained after applying both programmatic filtering \emph{i.e.} either the unit test fails (18.3\%) or returns the wrong answer (9.8\%), and text-based filtering ($\sim$ 50\% of the total from the previous stage). Note that we opt to filter out such a large percentage of QA pairs in the interest of ensuring high-quality evaluation data. Further, our benchmark curation process is fully automatic and so can be readily scaled to a larger image-caption source. Questions in \methodname~average 10.3 words in length whereas answers average 13.4 words (see Fig.~\ref{fig:dataset}, \emph{right}). In Fig.~\ref{fig:dataset}, \emph{left} we present a sunburst visualization of the first 4 words in the questions, highlighting the diversity of questions in our benchmark.

\subsection{Programmatic VLM Evaluation (\methodname)}
\label{sec:eval}

After ensuring the validity of the generated QA pairs, we proceed to evaluating free-form VLM responses to the same \response=\model(\question, \image). We first extract tuples from \response~(using an LLM~\citep{dubey2024llama} with in-context prompting), that we use to build a scene graph representation \scenegraph(\response). We also build a similar scene graph from the ground truth answer tuples after excluding ``premise'' tuples included in the question $\scenegraph(\answer)-\scenegraph(\question)$. We then measure response helpfulness \help(.) based on \emph{recall} of this scene graph, \emph{i.e.} the fraction of tuples (nodes, attributes, and relationships) in $\scenegraph(\answer)-\scenegraph(\question)$ that are recovered by \scenegraph(\response). 
Concretely, we compute average cosine similarity between each ground truth tuple and its closest response tuple in  embedding~\citep{reimers2019sentence} space. 
\begin{equation}
    \help(\response) = \frac{\sum_{t \in \scenegraph(\answer)-\scenegraph(\question) } \max_{t' \in \scenegraph(\response)} \text{sim}(t, t')}{|\scenegraph(\answer)-\scenegraph(\question)|};\;\;\;
    \label{eq:metrics_help}
\end{equation}

Next, we compute \truth(.) as the \emph{precision} of the response \emph{i.e.} the fraction of response tuples that are consistent with either the original scene graph \emph{or} the image itself\footnote{This reduces false-positive hallucination detections, as no caption can capture every aspect of an image.}. We define:
\begin{equation}
    \truth(\response) = \frac{\sum_{t' \in \scenegraph(\response)}  \text{max}\big(\max_{t \in \scenegraph(\imagecaption)}\text{sim}(t', t), p(\image \models t')\big) }{|\scenegraph(\response)|};\;\;\;
    \label{eq:metrics}
\end{equation}

where $\models$ denotes visual entailment, and $p(\image \models t')$ is approximated using a visual entailment model~\citep{cai2019once}. Note that \help~and \truth~are not necessarily correlated -- a response can be helpful (by answering the query) but not entirely truthful (might contain hallucinations), and vice versa. Naturally, different models may achieve different trade-offs between the two -- an aspect that \methodname~is uniquely suited to analyze.

%% file: sections/4_experiments.tex
\section{Experiments}
\label{sec:experiments}

\begin{table}[t]
    \centering
        \begin{tabular}{lcccc}
            \toprule   
            \textbf{Method} & \textbf{\#params} & \textbf{\help~($\uparrow$)} & \textbf{\truth~($\uparrow$)}  & \textbf{average~($\uparrow$)} \\
            \midrule[0.75pt]
            Qwen2-VL~\citep{bai2023qwen} & 2B & 69.36&80.64&75.00 \\            
            InternVL2~\citep{chen2024far} & 2B & \textbf{73.96}&79.51&76.74\\
            Phi-3.5-Vision~\citep{abdin2024phi} & 4B & 73.35&\textbf{82.27}&\textbf{77.81}\\
            \midrule[0.25pt]                        
            LLaVA-1.5~\citep{liu2024visual} &  7B & 72.67&\textcolor{blue}{\textbf{82.58}}&\textbf{77.62}\\
            LLaVA-Next~\citep{liu2024llava} &  7B& 74.28&80.03&77.15\\            
            InternVL2~\citep{chen2024far} & 8B & \textbf{74.55}&80.56&77.56\\                        
            \midrule[0.25pt]                                    
            Pixtral~\citep{pixtral12b} & 12B & 73.34 &\textbf{82.43}&\textbf{77.88}\\
            LLaVA-1.5~\citep{liu2024visual} &  13B & 72.46 & 82.40 & 77.43 \\                             
            InternVL2~\citep{chen2024far} & 26B & \textbf{74.63}&79.23&76.93 \\       
            \midrule[0.25pt]                        
            Claude-3.5-Sonnet$^\dagger$~\citep{claude32023} & - & 71.06&77.31&74.19 \\
            GPT-4o-mini$^\dagger$~\citep{achiam2023gpt} & - & 73.18&79.24&76.21 \\
            Gemini-1.5-Flash$^\dagger$~\citep{team2023gemini} &  -& 72.73&\textbf{81.74}&77.23 \\                                                            
            GPT-4o$^\dagger$~\citep{achiam2023gpt} & - & \textcolor{blue}{\textbf{76.53}} & 80.92 & \textcolor{blue}{\textbf{78.72}} \\
            \midrule[0.25pt]
            \textcolor{gray}{Oracle*} & - &\textcolor{gray}{82.84}&\textcolor{gray}{85.59}&\textcolor{gray}{84.22}\\
            \bottomrule
        \end{tabular}
    \caption{Benchmarking VLMs on \methodname~(*=LLaMA-3.1~\citep{dubey2024llama} backbone, $^\dagger$=closed-source). For each model, we report helpfulness (\help), truthfulness (\truth), and their average. We find larger and more recent models achieve higher \help~ but not necessarily higher \truth~.}
    \label{tab:results}
\end{table}

We now present our benchmarking experiments on \methodname. We include a broad set of models spanning a range of sizes and learning strategies and extensively analyze their performance, including their performance trade-offs. We also conduct a human study to validate both the quality of our benchmark and how well our proposed metrics correlate with human judgement. 

\subsection{Setup}
\label{sec:setup}

\noindent\textbf{Baselines.} We include VLMs of three sizes -- small (<5B parameters), medium (5-10B parameters), and large (>10B parameters) -- and include both open-source and proprietary models. We also benchmark an additional LLM-based ``oracle'' model as an upper bound (a blind model that is provided with the ground truth caption image). For the model, we use a LLaMA-3.1-8B backbone~\citep{dubey2024llama} with in-context prompting.

\noindent\textbf{Data.} \methodname~ is constructed from images, tuples, and captions released under a CC by 4.0 license as the test split of the DOCCI~\citep{onoe2024docci} dataset. DOCCI images were reviewed both by human and automatic methods to remove or obfuscate PII (faces, phone numbers, and URLs) and unsafe content. Images underwent a rigorous 3-stage human annotation phase resulting in hyper-detailed and high-recall captions averaging 136 words. 

\noindent\textbf{Implementation details.} We use GPT-4o~\citep{achiam2023gpt} for generating structured question, answers, and verification programs using the batch API and prompting it with a detailed task description, examples, and a Python definition of the SceneGraph class. We also use GPT-4o for the first round of text-based post-processing described in Sec.~\ref{sec:build}. We use OFA~\citep{cai2019once} fine-tuned for visual entailment for both post-processing and measuring image-tuple entailment (Eq.~\ref{eq:metrics}), and Sentence-Bert ~\citep{reimers2019sentence} to extract text embeddings. 
\begin{figure}[t]
    \centering
    \includegraphics*[width=0.57\linewidth]{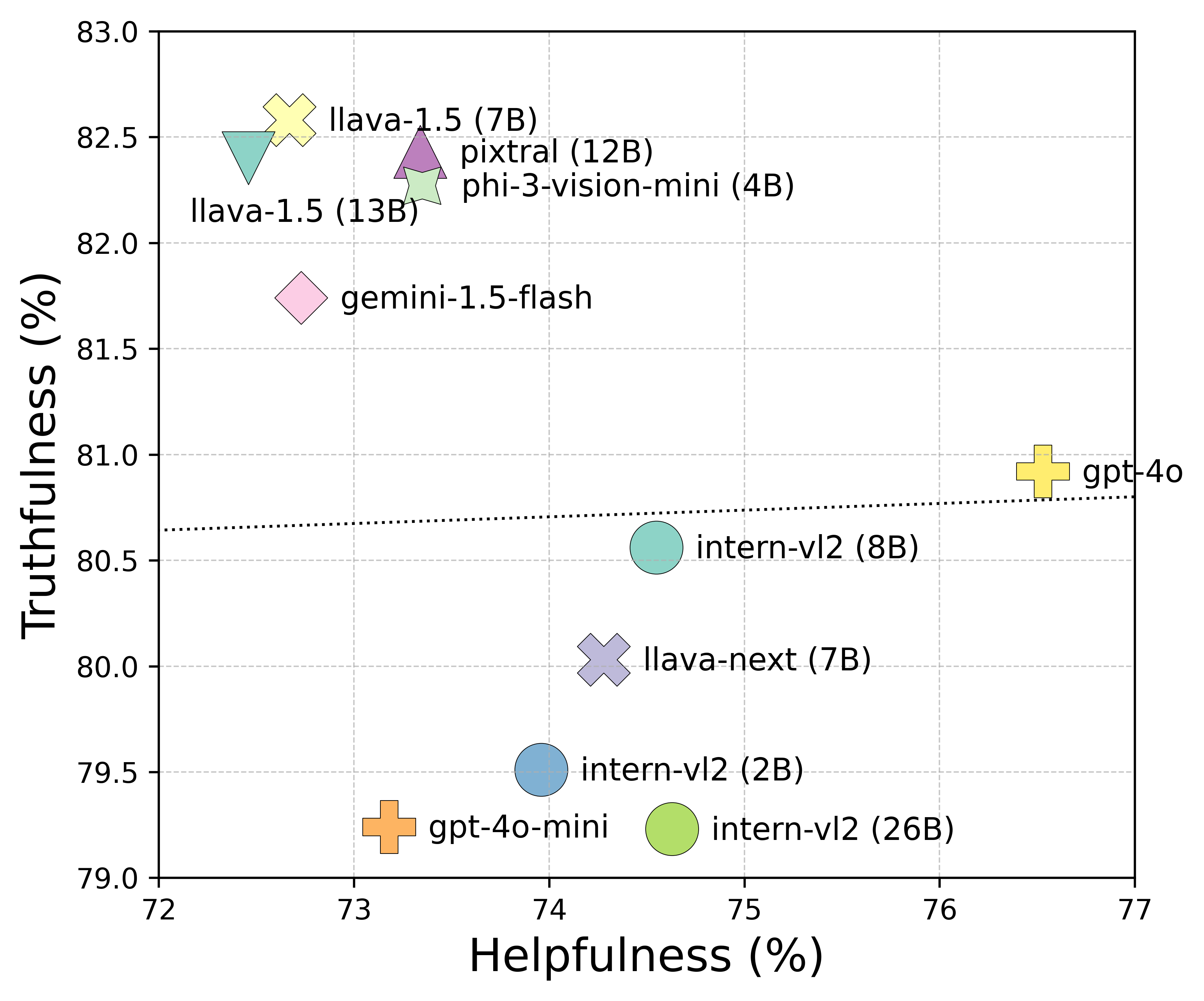}
\caption{\label{fig:tradeoff} We plot \help~and \truth~for VLMs on \methodname~-- as seen, models with higher helpfulness tend to lag behind on truthfulness, with very few striking a good trade-off between the two. Averaged across models, we observe a weak linear correlation of \textbf{0.03} between \help~and \truth~.
}
\end{figure}

\subsection{Results}
\label{sec:results}

Table~\ref{tab:results} and Figure~\ref{fig:tradeoff} present evaluation results. We find that:

\noindent\textbf{$\triangleright$ Few models strike a good balance between helpfulness and truthfulness.} As Fig.~\ref{fig:tradeoff} (left) shows, models tend to exhibit a range of trade-offs between helpfulness and truthfulness, with very few from the subset that we study (GPT-4o, Phi-3.5-Vision, Pixtral) managing to strike a good balance between the two. In fact, we find that many recent models that rank highly on perception and reasoning-focused aggregate benchmarks~\citep{lu2024wildvision}, such as Claude-3.5-Sonnet~\citep{claude32023} and Intern-VL2 (26B)~\citep{chen2024far} \emph{do not} necessarily translate to high truthfulness on \methodname, lagging behind simpler and smaller models like LLaVA-1.5~\citep{liu2024visual} in \truth. In fact, we find the LLaVA-1.5 model series to obtain the best \truth~overall. Overall, we observe a weak linear correlation of \textbf{0.03} between \help~and \truth~averaged across models, suggesting that the impressive recent gains in model helpfulness have not necessarily translated to higher truthfulness.

Table~\ref{tab:results} shows a more detailed breakdown of performance and includes additional baselines, categorized by parameter count. As seen, the oracle model
does considerably better than other models, indicating that there is still significant room for improvement in the current generation of VLMs.

\noindent\textbf{$\triangleright$ Increasing model size improves \help~ but not necessarily \truth.} Across both model families that we benchmark at multiple sizes -- InternVL2~\citep{chen2024far} (2B, 8B, and 26B), and LLaVA~\citep{liu2024llava} (1.5-7B, Next-7B, and 1.5-13B), we find that larger or more recent variants tend to outperform smaller ones in terms of helpfulness but not necessarily truthfulness.

\begin{figure}[t]
    \centering
    \includegraphics*[width=\linewidth]{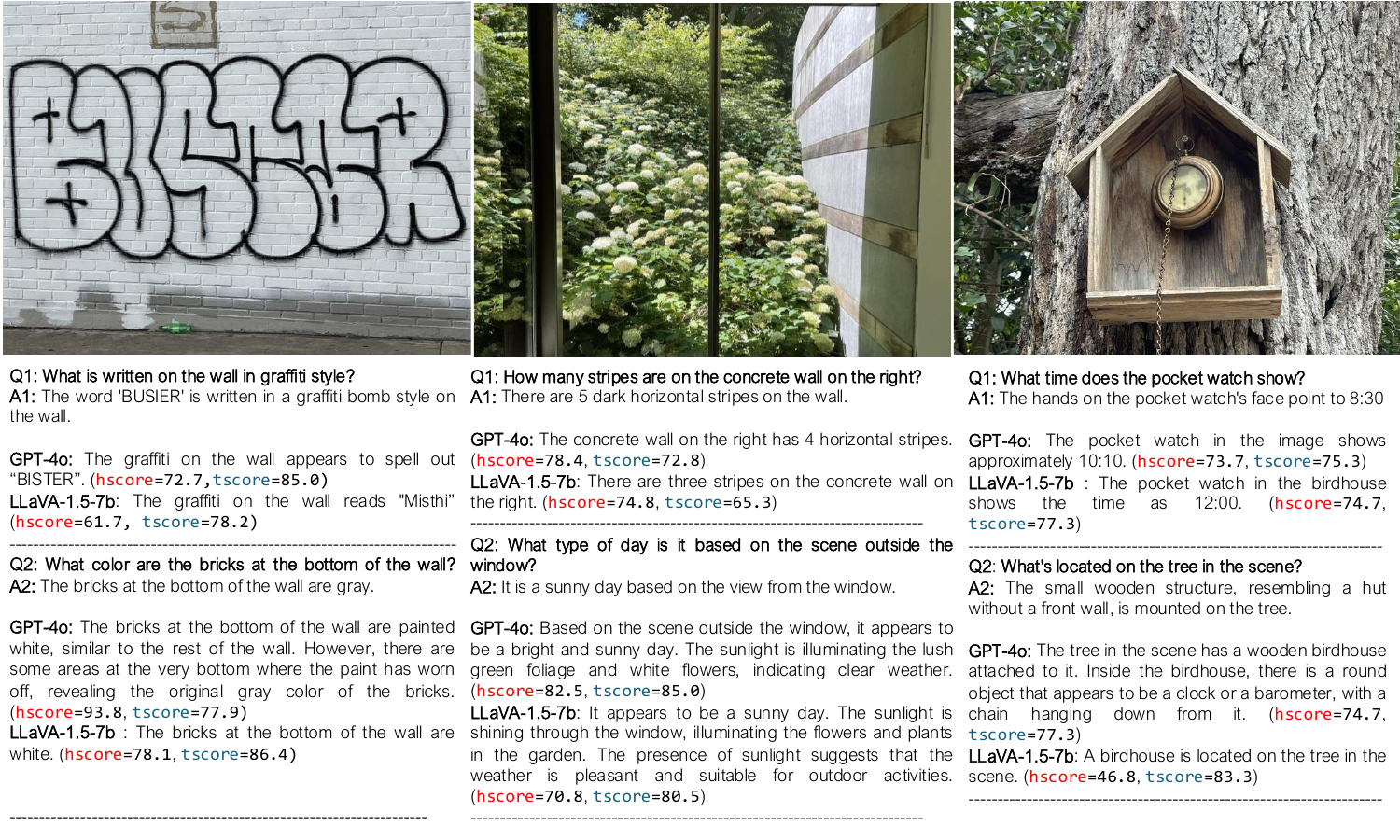}
    \caption{\label{fig:qual} Example responses from two VLMs that achieve high \help~(GPT-4o) and \truth~(LLaVA-1.5 (7B)) respectively. While both models struggle with sub-tasks such as OCR, counting, and reading an analog clock, GPT-4o's errors tend to be less egregious which leads to a higher \help.}
    \label{fig:qualitative}
\end{figure}

\noindent\textbf{$\triangleright$ Models fail in different ways.} In Fig.~\ref{fig:qual} we provide example responses from two models with high \help~(GPT-4o) and \truth~(LLaVA-1.5-7B) respectively. We find that while both models struggle with subtasks such as OCR, counting, and reading an analog clock, GPT-4o's errors tend to be less egregious (\emph{e.g.} reading 3/6 letters of the graffiti correctly, while LLaVA only gets 1/6). Further, GPT-4o tends to generate more descriptive answers (\emph{e.g.} correctly identifying that while the wall in the first image is white, the bricks at the bottom are gray), which boost its \help. In Fig.~\ref{fig:scores}, we include a fine-grained analysis of GPT-4o performance across different question types.

\noindent\textbf{Human evaluation of \methodname~ and proposed metrics.} Finally, we conduct two human studies of our benchmark. We first ask human annotators (3 per example) to evaluate the question relevance and answer correctness of QA pairs generated from the qual-test split of the DOCCI dataset (100 images, 170 generated QA pairs) that is specifically set aside for human evaluation. After majority voting, annotators judge 163/170 questions to be relevant (95.9\%) and 167/170 answers to be correct (98.2\%). We manually inspect the small number of examples judged as irrelevant or incorrect in and find most to be either particularly challenging or subjective, rather than irrelevant or incorrect.

In the second study, we ask subjects (3 per example) to \emph{rate} responses from four models -- GPT-4o, LLaVA-1.5-7B, LLaVA-Next-7B, and GPT-4o-mini -- on the same set of 170 QA pairs based on their helpfulness (0=unhelpful, 1=helpful) and truthfulness (0=fully false, 0.5=partially false, 1.0=fully true), and average. We then automatically compute \help~and \truth~for the same set of responses and measure the Pearson correlation between the two, observing a strong correlation of 0.81 for \help~and a modest correlation of 0.45 for \truth, supporting the validity of these metrics.

%% file: sections/5_discussion.tex
\section{Discussion}
\label{sec:discussion}

Our work takes a step towards reliably evaluating the helpfulness-truthfulness trade-offs of vision-language models. Our design leverages an LLM prompted with a robust scene graph representation and API to construct ``in-the-wild'' visual question answer pairs that are \emph{grounded by design}. Further, these QA pairs lend themselves to programmatic evaluation via comparing scene-graph representations. The reliability of our benchmark comes from three factors: i) high-recall human-annotated image captions that seed the scene graphs, which make it possible to (almost) exhaustively validate the veracity of any claim ii) programmatic verification of the generated QA pairs that ensure that both the question and answer are indeed grounded in the visual input, and iii) evaluation metrics that are both holistic (\emph{i.e.} consider all the provided context) and interpretable (\emph{i.e.} provide a concrete scoring rubric based on scene graph-based matching). 

\noindent\textbf{Limitations.} While we hope that \methodname~will serve as a useful test-bed for reliable VLM evaluation and spur future research on the topic, it is not without limitations. While we try to ensure high precision in QA pairs retained in our benchmark (via programmatic verification), this naturally comes at some cost to recall (\emph{i.e.} some hard-to-verify question types may be excluded from the benchmark). Next, even high-recall image captions may not capture every aspect of an image, and so our evaluation may not be able to catch all model hallucinations. Further, our evaluation relies on off-the-shelf models for computing text-embeddings, scene graph tuples, and image-text entailment, and so almost certainly inherits some of their limitations. Finally, we hope future work will study the effectiveness of recent fine-tuning~\citep{liu2023mitigating}, preference-tuning~\citep{yu2024rlhf,sun2023aligning}, and training-free~\citep{huang2024opera,kim2024if,leng2024mitigating} hallucination mitigation strategies on \methodname, as well as  agentic models that can plan~\citep{suris2023vipergpt,gupta2023visual}, reason, and self-reflect~\citep{valmeekam2024llms}, towards the elusive goal of achieving Pareto improvements in both helpfulness and truthfulness.

\noindent\textbf{Acknowledgements.} We would like to thank Jieyu Zhang and Artemis Panagopoulou for valuable feedback and discussions, Juan-Carlos Niebles, Gabriel Bernadette Shapiro, and Silvio Savarese for feedback on our draft, and Srinath Reddy Meadusani for help with compute infrastructure.

%% file: sections/X_suppl.tex
\definecolor{codegreen}{rgb}{0,0.6,0}
\definecolor{codegray}{rgb}{0.5,0.5,0.5}
\definecolor{codepurple}{rgb}{0.58,0,0.82}
\definecolor{backcolour}{rgb}{0.95,0.95,0.92}

\lstdefinestyle{mystyle}{
    backgroundcolor=\color{backcolour},   
    commentstyle=\color{codegreen},
    keywordstyle=\color{magenta},
    numberstyle=\tiny\color{codegray},
    stringstyle=\color{codepurple},
    basicstyle=\ttfamily\footnotesize,
    breakatwhitespace=false,         
    breaklines=true,                 
    captionpos=b,                    
    keepspaces=true,                 
    numbers=left,                    
    numbersep=5pt,                  
    showspaces=false,                
    showstringspaces=false,
    showtabs=false,                  
    tabsize=2
}

\lstset{style=mystyle}

\begin{figure}
    \centering
        \centering
        \begin{subfigure}[t]{0.45\columnwidth}
            \centering
        \includegraphics*[width=\linewidth]{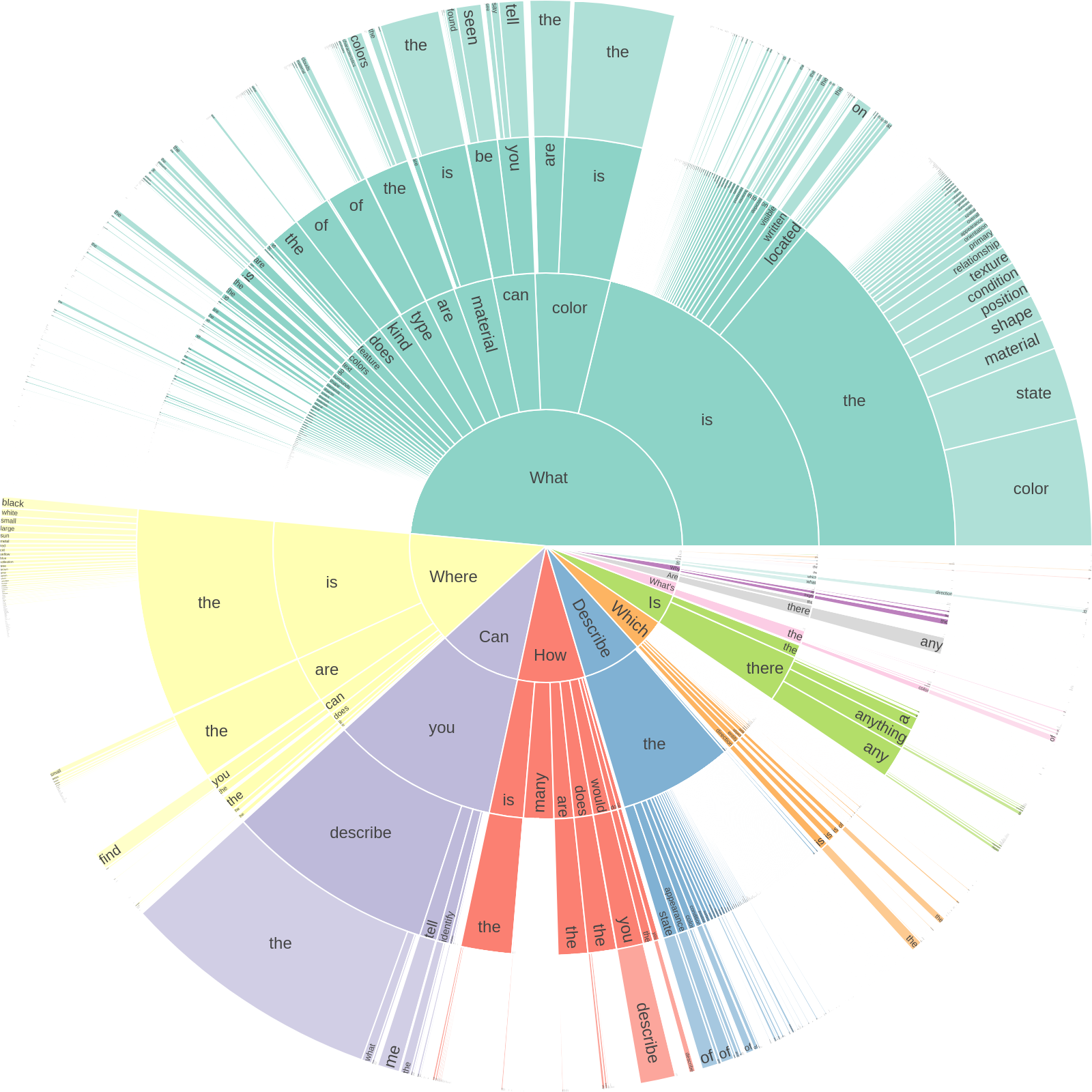}
        \caption{\label{fig:sunburst} Sunburst visualization of the first 4 question words in ~\methodname.}        
        \end{subfigure}
        \begin{subfigure}[b]{0.45\columnwidth}
            \centering
            \includegraphics*[width=\linewidth]{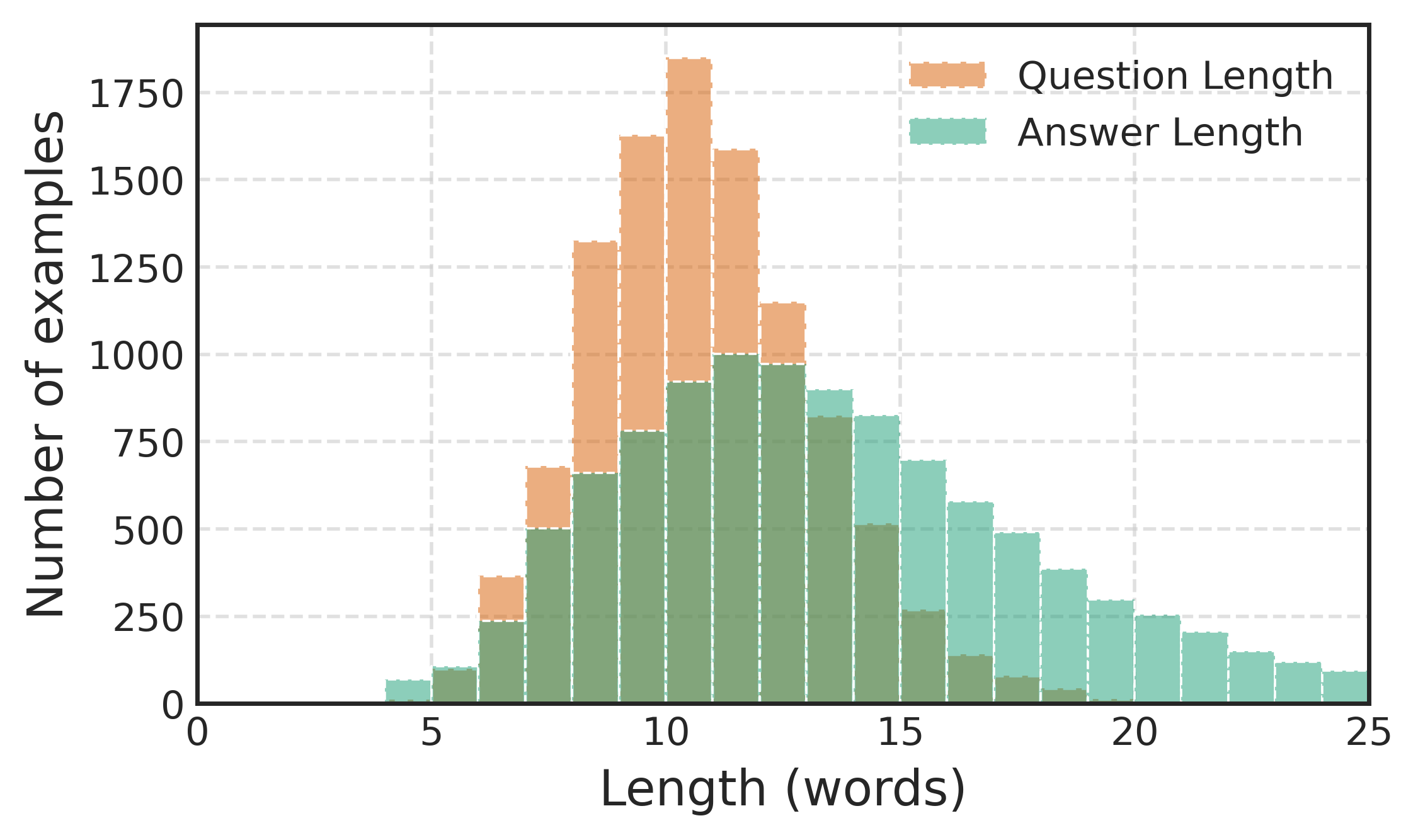}
        \caption{\label{fig:hist} Distribution of question and answer lengths in \methodname.}
        \end{subfigure}
        \caption{\label{fig:dataset} \methodname: Additional dataset statistics.}
\end{figure}

\subsection{Additional dataset details}

Fig.~\ref{fig:sunburst}, \emph{left} presents a sunburst visualization of the first 4 words in the questions within the \methodname~dataset. As seen, the questions are diverse and span a wide range of question types. Further, while nearly 50\% of the questions begin with ``What'', even this subset spans a range of topics testing numerous model capabilities -- see Fig.~\ref{fig:qualitative}. Fig.~\ref{fig:dataset}, \emph{right} shows the distribution of question and answer lengths in \methodname. Questions in the dataset average 10.3 words in length, whereas answers average 13.4 words, with both following a normal distribution spread. 

\subsection{Additional implementation details}

Lst.~\ref{lst:python} provides a Python implementation of the SceneGraph class used to represent scene graphs in \methodname. The class provides methods to generate subgraphs, describe subgraphs in natural language, and query entities, attributes, and relationships. We include the prompts used for generating verifiable question-answer pairs as well as the post-processing prompt used to filter out low-quality QA pairs in Figs.~\ref{fig:prompt1}-~\ref{fig:prompt2}. 

\subsection{Additional performance analysis}

\begin{figure}[t]
    \centering
    \begin{subfigure}[b]{0.5\columnwidth}
        \includegraphics*[width=\linewidth]{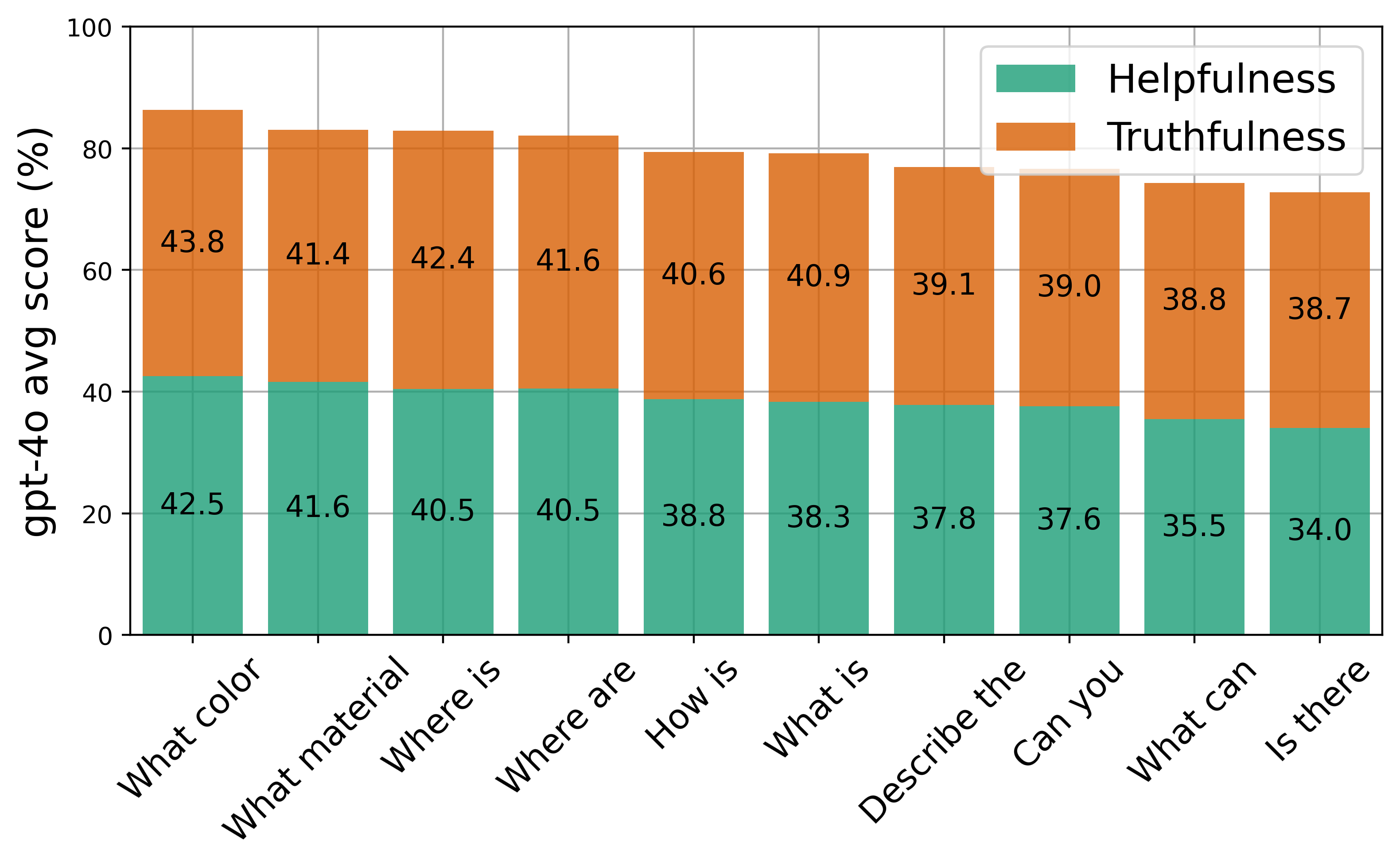}
    \caption{\label{fig:scores} GPT-4o fine-grained performance analysis}
    \end{subfigure}
    \begin{subfigure}[b]{0.42\columnwidth}
    \includegraphics*[width=\linewidth]{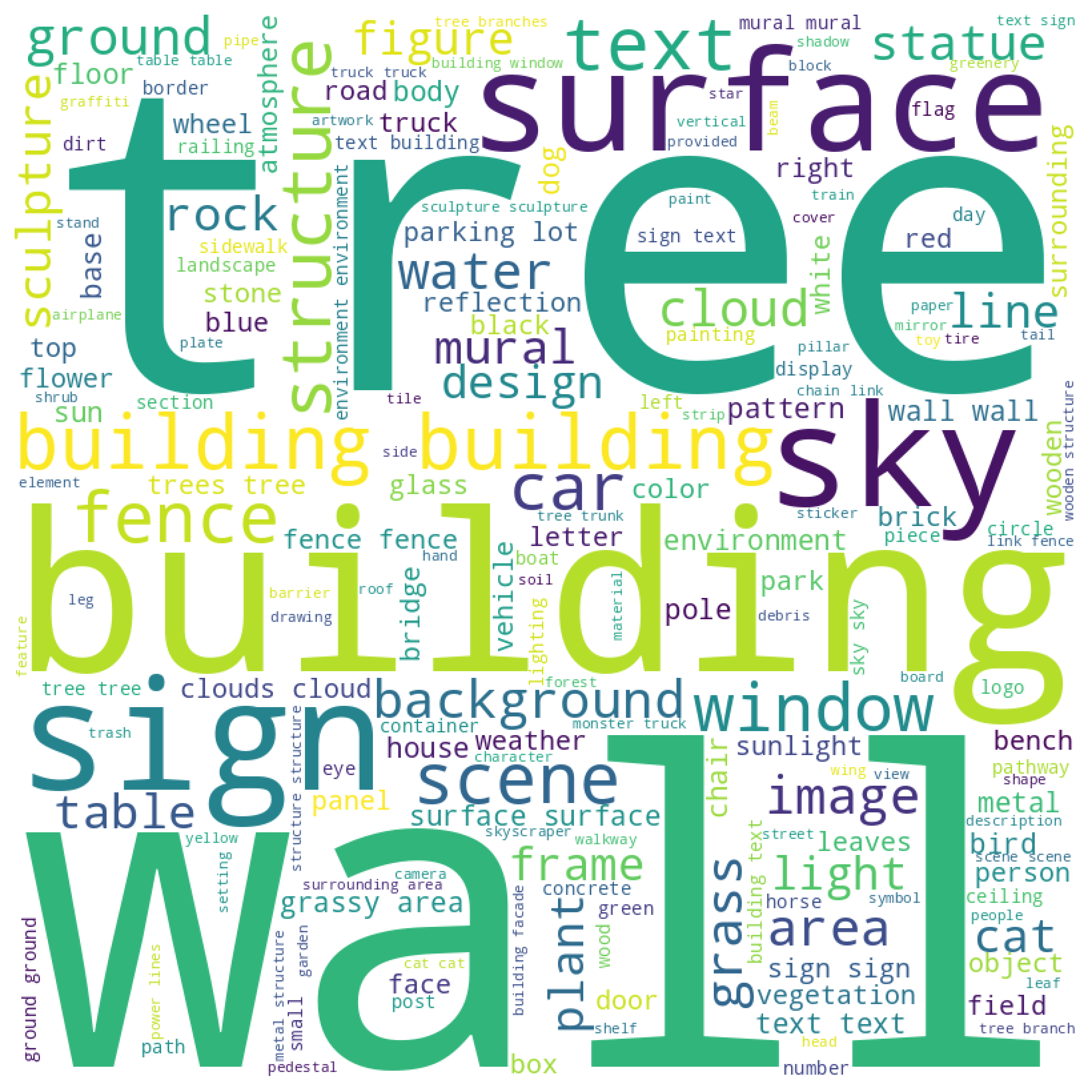}
    \caption{\label{fig:hal} Word cloud of hallucinated objects in model-generated responses.}
    \end{subfigure}
    \caption{\label{fig:fg} Fine-grained performance analysis.}
\end{figure}

Fig.~\ref{fig:scores} presents a fine-grained performance analysis of GPT-4o on \methodname. We break down helpfulness and truthfulness scores by question type, and display the top-10 most common question types sorted by performance. As seen, the model performs particularly well on questions that require reasoning about spatial relationships (where are/is), object attributes (what color), and generating image descriptions.

Fig.~\ref{fig:hal} shows a word cloud of the most commonly hallucinated objects in answers to questions from \methodname~across all models. As seen, models commonly hallucinate common objects such as ``tree'', ``building', ``wall'', and ``sign''. 
     
\clearpage
\begin{lstlisting}[language=Python, caption=Python API for the SceneGraph class, label=lst:python]
########################################################
# SceneGraph class API
########################################################
class SceneGraph(nx.DiGraph):
    def __init__(self, caption, sg_dict, *args, **kwargs):
        """Init scene graph from entity-attribute-relationship dict"""
        super().__init__(*args, **kwargs)
        for source_ent, metadata in sg_dict.items():
            self.add_node(source_ent, **metadata["attributes"])
            for target_ent, rel_info in metadata["relations_to"].items():
                self.add_edge(source_ent, target_ent, **rel_info)
        self.caption = caption
        self.sg_dict = sg_dict

    def generate_subgraph(self, node_list)->"SceneGraph":
        """Generates a subgraph with nodes in node_list"""
        return nx.subgraph(self, node_list)
        
    def describe(self, subgraph): -> str:
        """Generate a natural language description of a subgraph"""
        return generate_description(subgraph)

    def get_entities(self) -> List[str]:
        """Returns a list of entities in the scene graph."""
        return list(self.nodes)

    def get_attributes(self, ent_name) -> dict[List]:
        """
        Returns a list of attributes for ent_name in the scene graph
        Format: { "att_type": f"att_value_1,  att_value_2, ..." }
        """
        return self.nodes.get(ent_name, {})

    def get_outgoing_relations(self, ent_name) -> dict:
        """
        Returns a dict of relations for which ent_name is the source.
        Format:  {target_ent_1: { rel_type_1: [rel_val_1, ...]} ... }
        """
        out_edges = list(self.out_edges(ent_name, data=True))
        out_edges = { tup[1]: {**tup[2]} for tup in out_edges }
        return out_edges
    
    def get_incoming_relations(self, ent_name) -> dict:
        """
        Returns a dict of relations for which ent_name is the target 
        Format: {source_ent_1: { rel_type_1: [rel_val_1, ...]} ...}
        """                    
        in_edges = list(self.in_edges(ent_name, data=True))
        in_edges =  { tup[0]: {**tup[2]} for tup in in_edges }
        return in_edges

    \end{lstlisting}

\clearpage

\begin{figure}
\noindent\fbox{%

    \parbox{\textwidth}{%
    Consider the SceneGraph class defined below, which takes as input an image caption and a dictionary of tuples (entities, attributes, and relations) parsed from the image caption, and builds a directed graph representation with attributed entities as nodes and relations as edges.\\

    You will be provided with such an image caption, tuple dictionary, and corresponding SceneGraph object. Your task is to generate a set of: 
    \begin{enumerate}
        \item \textbf{Free-form question-answer pairs} that test non-trivial image understanding and reasoning capabilities.
        \item \textbf{A Python function} that receives as input the SceneGraph object and can be executed to answer the query by reasoning over the scene graph.
    \end{enumerate}
    
    \textbf{Guidelines.} The generated questions should be:
    \begin{itemize} 
    
        \item \textbf{Clear and conversational}, in the tone of a person who is asking another person about the scene. You may paraphrase where appropriate to improve clarity (eg. ``Can you describe the dog?'' is better than ``What is the state of the dog?''). Note that the tuples in the scene graph are generally accurate but not necessarily precise, and so may require rephrasing to generate meaningful questions from.
    
    \item \textbf{Diverse}, both in question type (\emph{e.g.} starting with ``is'', ``where'', ``what'', ``when'', ``how'', ``which'', ``why'', etc.) and length.
    
    \item \textbf{Non-trivial} (eg. avoid ``What color are the green trees?'') and \textbf{unambiguous} (eg. avoid ``What is the color of the puppy?'' for an image with multiple puppies).
\end{itemize}      
    
The generated Python functions should be:
\begin{itemize}      
    \item \textbf{Executable}: The code should run without requiring modifications.
    
    \item \textbf{General}: The code should generalize to similar scene graphs as the one provided. Do NOT hard-code specific attributes or relations.
\end{itemize}   
    For each image, generate 10-15 such question-answer pairs and corresponding Python functions. 
    }
}
\caption{\label{fig:prompt1} LLM prompt for generating visual question-answer pairs along-with verification programs.}
\end{figure}

\clearpage
\begin{figure}
\noindent\fbox{%

\parbox{\textwidth}{%
You will be provided with a list of question-answer pairs about an image. Your task is to identify whether each pair has any of the following issues:
\begin{enumerate}

    \item  \textbf{Trivial question}. A trivial question can be answered directly from information provided in the question or using common-sense, without requiring looking at the image. Examples:

    \textbf{Question}: What is the material of the stadium's horizontal concrete bar?\\
    \textbf{Answer}: The stadium's horizontal concrete bar is made of concrete.\\
    \textbf{Judgement}: Trivial (The question already mentions that the bar is made of concrete)\\

    \textbf{Question}: What text rendering is found on the stop sign?\\
    \textbf{Answer}: The stop sign has white text rendering of the word "STOP".\\
    \textbf{Judgement}: Trivial (Stop signs almost always have the word "STOP" on them)\\

    \textbf{Question}: What feature of the scene reflects sunlight?\\
    \textbf{Answer}: The hard surfaces reflect sunlight.\\
    \textbf{Judgement}: Trivial (Hard surfaces are known to reflect sunlight)\\

    \item  \textbf{Incomplete answer.} An incomplete answer does not completely answer the question. It may be missing key details or may not provide a full description, or may also be entirely irrelevant. Examples:

    \textbf{Question}: What is between the red neon light and the frame?\\
    \textbf{Answer}: The red neon light is behind the metal construction frame.\\
    \textbf{Judgement}: Incomplete (does not answer the question)\\

    \textbf{Question}: How would you describe the trees surrounding the green lake?\\
    \textbf{Answer}: The trees surrounding the green lake are large in size.\\
    \textbf{Judgement}: Incomplete (``large'' is not a sufficiently detailed description)\\

    \textbf{Question}: In which part of the image is there no visible cloud coverage?\\
    \textbf{Answer}: The rest of the image has the clear blue sky with no visible cloud coverage.\\
    \textbf{Judgement}: Incomplete (``the rest of the image'' is meaningless without context)\\

    \item  \textbf{Unnatural-sounding}. The question-answer pair may sound awkward, ambiguous, or unnatural. This could be due to its phrasing, structure, or grammar. Examples:

    \textbf{Question}: Can you describe the role of the stones in relation to the anemones?\\
    \textbf{Answer}: The stones are covered with anemones and line the bottom of the tank and go up its left side.\\
    \textbf{Judgement}: Unnatural (``role'' of stones is odd phrasing and the overall question is ambiguous)\\

    \textbf{Question}: What is the overall shape of the section of grass?\\
    \textbf{Answer}: The section of grass is small in shape.\\
    \textbf{Judgement}: Unnatural (``small'' is not a shape)\\

    \textbf{Question}: What kind of state is the sign experiencing due to the brightness?\\
    \textbf{Answer}: Due to the brightness, the sign is experiencing a duller state.\\
    \textbf{Judgement}: Unnatural (``experiencing a state'' is awkward phrasing)\\

\end{enumerate}
}}
\caption{\label{fig:prompt2}LLM text-based post-processing prompt. }
\end{figure}